\documentclass{article}

\usepackage{arxiv}

\usepackage{amsthm, amssymb, mathtools}
\newtheorem{thm}{Theorem}
\usepackage[]{algorithm2e}
\usepackage{dsfont}
\usepackage{xcolor}

\newcommand{\Adam}{\texttt{Adam}\xspace}
\newcommand{\AdamMCMC}{\texttt{AdamMCMC}\xspace}

\newcommand{\Var}{\mathrm{Var}}
\newcommand{\R}{\mathds{R}}
\newcommand{\train}{\mathcal{D}}

\newcommand{\temp}{\lambda}

\newcommand{\1}{\mathbf{1}}

\def\secref#1{Section~\ref{#1}}

\def\algref#1{Algorithm~\ref{#1}}

\usepackage{fancyhdr}
\usepackage[round]{natbib}
\usepackage{hyperref}

\setcitestyle{authoryear,round,citesep={;},aysep={,},yysep={;}}

\newcommand{\E}{\mathbb{E}}
\renewcommand{\theta}{\vartheta}
\renewcommand{\phi}{\varphi}

\allowdisplaybreaks

\title{\AdamMCMC: Combining Metropolis Adjusted Langevin with Momentum-based Optimization}

\usepackage{authblk}

\author[1]{Sebastian Bieringer\thanks{\texttt{sebastian.guido.bieringer@uni-hamburg.de}}}%
\author[1]{Gregor Kasieczka}
\author[2]{Maximilian F. Steffen}
\author[2]{Mathias Trabs}

\affil[1]{Universit\"at Hamburg, Institut f\"ur Experimentalphysik, Luruper Chaussee 149, 22761 Hamburg, Germany}
\affil[2]{Karlsruhe Institute of Technology, Institut f\"ur Stochastik, Englerstr. 2, 76131 Karlsruhe, Germany}

\renewcommand{\shorttitle}{AdamMCMC}

\hypersetup{
pdftitle={AdamMCMC},
pdfauthor={Sebastian Bieringer, Gregor Kasieczka, Maximilian F.~Steffen, Mathias Trabs},
}

\begin{document}

\maketitle

\begin{abstract}
Uncertainty estimation is a key issue when considering the application of deep neural network methods in science and engineering. 
In this work, we introduce a novel algorithm that quantifies epistemic uncertainty via Monte Carlo sampling from a tempered posterior distribution.
It combines the well established Metropolis Adjusted Langevin Algorithm (MALA) with momentum-based optimization using \Adam and leverages a prolate proposal distribution, to efficiently draw from the posterior.
We prove that the constructed chain admits the Gibbs posterior as invariant distribution and approximates this posterior in total variation distance. Furthermore, we demonstrate the efficiency of the resulting algorithm and the merit of the proposed changes on a state-of-the-art classifier from high-energy particle physics.
\end{abstract}

\section{Introduction}

Deep learning methods are widely applied in industry, engineering, science and medicine.
Especially in the latter fields, widespread application of such methods is held back by non-existent, overconfident or time-intensive error estimates.
This is especially problematic when neural network outputs are used 
in clinical decision making, 
control systems of autonomous vehicles, 
or scientific discovery, for example in particle physics~\citep{Karagiorgi:2022qnh}.

In these cases, the uncertainty in the data generation or taking, the \textit{aleatoric} uncertainty, can by accessed by learning a data likelihood $p(\train|\vartheta)$ in the same framework.
This is often done by parameterizing the likelihood as a Gaussian \citep{Gal2016Uncertainty} or with a normalizing flow architecture~\citep{NF_cond_radev2020bayesflow}.

Bayesian neural networks (BNNs) can be used to estimate an uncertainty on the neural network fit stemming from the limited training statistics, that is the \textit{epistemic} uncertainty.
Such methods understand the network parameters $\vartheta$ as random variables from a posterior distribution $p(\vartheta|\train)$ conditioned on the training data $\train$.
By drawing from this distribution we can produce an uncertainty prediction from an ensemble of different parameter samples.


BNN algorithms based on parametric estimates of the posterior weights, be it Gaussian-mean field~\citep{bbb2015} or Laplace approximation based~\citep{DaxbergerKIEBH21_laplace_redux, ritter2018kfac}, have to balance the quality of the fit with the complexity of the algorithm. 
To accommodate efficient evaluation and scaling with the network size, they rely on (block-) diagonal approximations of the covariance matrix, leading to bad fit performance and underconfident uncertainties.

Markov Chain Monte Carlo (MCMC) algorithms on the other hand can access the full weight posterior, but struggle with slow convergence and high computational costs.
To reduce computation, \citet{Welling2011bayesianSGLD} proposed a first chain based on stochastic gradients.
The authors achieve unbiased sampling in the limit of diminishing step sizes, thus reducing the phase space exploration of the algorithm to a random walk.
\Citet{chen2014stochastic} therefore adapt Hamiltonian Monte Carlo for stochastic gradients (sgHMC) to achieve more efficient exploration through use of an auxiliary momentum variable.
Thermostats \cite{ding2014_thermostats,Heek2019_ATMC_thermostats} use yet another auxiliary variable, inspired by thermodynamics, that is quadratic in the parameters to further improve the sampling.

The random walk like diffusion process underlying these stochastic gradient based chains can be tuned to the geometry of the posterior with a diffusion matrix~\citep{Patterson2013_fisher_sgLF,XIFARA2014_fisher}.
\Citet{ma2015complete} note that arbitrary, positive definite diffusion matrices can be used whenever the commonly used Fisher metric is hard to compute.
This includes the preconditioning of RMSprop~\citep{Li2016SGLD, Chen2016santa}.

Rather than using the general framework of \citet{ma2015complete} to construct a chain that admits the posterior as invariant distribution, we bridge the gap to stochastic optimization by
\begin{itemize}
    \item proposing a novel algorithm that employs the MALA algorithm~\citep{roberts1996b} around \Adam updates \citep{Adam2014}, including first and second order momentum.
    In combination with a prolate deformation of the diffusion we ensure high acceptance rates.
    We refer to the resulting algorithm as \AdamMCMC.

    \item For a state-of-the-art particle physics application, we show the changes lead to a more well-behaved algorithm, that approaches the performance of pure \Adam for narrow proposal distributions and allows adjusting the calibration as a tradeoff between fit uncertainty and performance.
\end{itemize}
\AdamMCMC uses a Metropolis-Hastings (M-H) step~\cite{Robert2004MCMC} that ensures convergence for fixed size learning rates.
It has desirable theoretical properties in terms of contraction rates as well as credible sets \citep{franssen2022, bieringer2023statistical}.
Our study can thus also be understood as a study of the difference between stochastic optimization and Bayesian inference in terms of M-H acceptance rates.

In~\secref{sec:MBB}, we introduce the established MALA algorithm. 
The changes to the proposal distribution are presented in~\secref{sec:aug}, with the accompanying proofs in Appendix~\ref{sec:proofs}.
We employ the proposed algorithm on an application from particle physics in Section~\ref{sec:example}, where we use a stochastic approximation of the M-H step. 
We discuss the implications of this choice, as well as proposals from related literature, in Section~\ref{ssec:stochMH} before concluding in~\secref{sec:conc}.

\section{Metropolis Adjusted Langevin Algorithm} \label{sec:MBB}

For a labeled or unlabeled $n$-point dataset $\train_n$, the $P$-dimensional vector of network weights $\vartheta$, a generic loss function $L(\vartheta)$ and its empirical counterpart $L_n(\vartheta)~=~L(\vartheta; \train_n)$, the Gibbs posterior is given by the density \begin{equation}
\label{eq:gibbs}
    p_\temp (\vartheta|\train_n) \propto \exp(-\temp L_n(\vartheta))~p(\vartheta),
\end{equation}
where $\temp>0$ is the inverse temperature parameter and $p(\vartheta)$ is a prior density on the network weights. The Gibbs posterior is a central object in the PAC-Bayes theory \citep{alquier2021} and it matches the classical Bayesian posterior distribution by Bayes theorem if $\lambda L_n(\vartheta)$ is the negative $\log$-likelihood of the data-generating distribution, that is $p(\mathcal D_n|\vartheta)=\exp(-\lambda L_n(\vartheta))$.
Throughout, we choose a uniform prior on some bounded set $\Omega\subset\mathbb R^P$.

Starting at some initial choice $\vartheta^{(0)}$, for every update step $k+1$ a new parameter proposal $\tau^{(k)}$ is sampled from a Gaussian \emph{proposal density} centered in a  gradient step
\begin{equation}
    \begin{aligned}
        \tau^{(k)} 
        \sim q(\tau|\vartheta^{(k)}) = 
        \frac{1}{(2\pi \sigma^2)^{P/2}}\exp \left( -\frac{1}{2\sigma^2}\left| \tau - \vartheta^{(k)} + \gamma \nabla_\vartheta L_n({\vartheta^{(k)}})\right|^2\right).
    \end{aligned}\label{eq:proposal}
\end{equation}
The new proposal is accepted with the \emph{acceptance probability} (ratio)
\begin{equation}
    \label{eq:acc_prob}
    \begin{aligned}
        \alpha(\tau^{(k)}| \vartheta^{(k)}) =
        \Bigg(\exp \left( -\temp L_n(\tau^{(k)}) + \temp L_n(\vartheta^{(k)}) \right) \cdot
        \mathbf{1}_\Omega(\tau^{(k)})\frac{q(\vartheta^{(k)}|\tau^{(k)})}{q(\tau^{(k)}|\vartheta^{(k)})}\Bigg)\wedge 1,
    \end{aligned}
\end{equation}
where $a\wedge b= \min\{a,b\}$.
Here, $q(\vartheta^{(k)}|\tau^{(k)})$ is the probability of the current parameters given the parameter proposal, that is the \textit{backwards direction}.
When the gradient steps are larger than $\sigma$, this probability can be very small and lead to vanishing acceptance probabilities.

If the proposal is not accepted, the previous parameter values are kept:
\begin{equation*}
    \vartheta^{(k+1)} = 
    \left\{
    \begin{array}{ll}
        \tau^{(k)} \text{ with probability } \alpha(\tau^{(k)}| \vartheta^{(k)}) \\
        \vartheta^{(k)} \text{ with probability } 1-\alpha(\tau^{(k)}| \vartheta^{(k)}).
    \end{array}
\right.
\end{equation*}
The calculation of the acceptance probability requires the evaluation of the loss function $L_n$ and its gradient $\nabla_{\vartheta} L_n$ for both the current weights $\vartheta^{(k)}$ and the proposed update $\tau^{(k)}$.

The choice of proposal density ensures that for a proposal $\tau^{(k)}$ close to the previous parameters $\vartheta^{(k)}$, the transition probability $q(\tau^{(k)}|\vartheta^{(k)})$ is bounded from below. 
Under these assumptions, \citet[Theorem 2.2]{10.1093/biomet/83.1.95} prove that $(\vartheta^{(k)})_{k \in \mathds{N}_0}$ is a Markov chain with invariant distribution $p_\temp (\cdot |\train_n)$. The resulting neural network estimators are denoted by $\hat f_{\vartheta^{(k)}}$.

After a \emph{burn-in} time $b \in \mathds{N}$, the Markov chain stabilizes at its invariant distribution and generates samples from the Gibbs posterior \eqref{eq:gibbs} in every iteration. 
To ensure approximate independence of the samples, we estimate the posterior mean prediction 
over samples with a gap of \emph{gap length} $c \in \mathds{N}$
\begin{equation*}
    \bar{f}_\lambda = \frac{1}{N}\sum_{i=1}^N \hat{f}_{\vartheta^{(b+ic)}}.
\end{equation*}

\section{Improved Sampling Efficiency} \label{sec:aug}


Proposals unlikely under the Gibbs posterior will come with low acceptance probabilities.
We thus need to make sure the proposals track the loss landscape closely as not to slow down the algorithm.
For fast convergence, it is thus desirable to run MALA at low $\sigma$.
This, on the contrary, reduces the probability of the backwards direction.
To solve this dilemma, we adapt the diffusion process and introduce a prolate proposal distribution.
In combination with momentum-smoothed trajectories (see Section~\ref{sec:math}), we recover high probabilities of the backwards direction even at large $\sigma$.

\subsection{Directional Noise}

Similarly to~\citet{ludkin2023hug_and_hop}, we replace the isotropic noise in \eqref{eq:proposal} with a modified Langevin proposal with a \emph{directional noise} in gradient direction $\nabla L_n^{(k)}:=\nabla_\vartheta L_n(\vartheta^{(k)})$. 
The new proposal distribution
\begin{equation}
    \begin{aligned}
        \tau^{(k)} \sim 
        q(\tau|\vartheta^{(k)}) =
        \frac{1}{\sqrt{(2\pi)^{P} \det(\mathbf{\Sigma}_k)}}  \exp \left( -\frac{1}{2} \left(\tau - \tilde{\vartheta}^{(k+1)}\right)^\top \mathbf{\Sigma}_k^{-1} \left(\tau - \tilde{\vartheta}^{(k+1)}\right) \right),
    \end{aligned}\label{eq:proposal2}
\end{equation}
is centered in
\begin{equation*}
    \tilde{\vartheta}^{(k+1)} = \vartheta^{(k)}+ \gamma \nabla_\vartheta L_n({\vartheta^{(k)}},
\end{equation*}
with covariance 
\begin{equation*}
    \mathbf{\Sigma}_k :=
  \mathbf{\Sigma}\left(\nabla L_n^{(k)}\right) := \sigma^2 I_{P}+\sigma^2_\nabla \nabla L_n^{(k)}\big(\nabla L_n^{(k)}\big)^\top.
\end{equation*}
Here, $\sigma,\sigma_\nabla>0$ are noise levels and $I_{P}$ the $P$-dimensional unit matrix. 
We omit the argument and write $\mathbf{\Sigma}_k$ whenever the direction of the update is clear from context.
In particular, the variance of $\tau^{(k)}$ in gradient direction is 
\[\Var\big(\langle \tau^{(k)},\nabla L_n^{(k)}\rangle\big)=\sigma^2|\nabla L_n^{(k)}|^2+\sigma_\nabla^2|\nabla L_n^{(k)}|^4,\]
while the variance in any direction $v\in\R^P$ orthogonal to $\nabla L_n^{(k)}$ is $\Var(\langle \tau^{(k)},v \rangle)=\sigma^2|v|^2$. 
As a consequence, the underlying random walk converges more reliably towards a minimum of the loss while still being sufficiently randomized to explore the whole parameter space. 

To evaluate the acceptance probability under the new proposal distribution effectively, we need to circumvent matrix multiplications in the calculation of both the determinant and inverse of the covariance $\mathbf{\Sigma}_k$. Thanks to our particular choice of the anisotropic covariance structure, 
the Sylvester determinant identity yields
\begin{equation}\label{eq:det}
\det(\mathbf{\Sigma}_k)=\sigma^{2P}\Big(1+\frac{\sigma^2_\nabla}{\sigma^2}\left|\nabla L_n^{(k)}\right|^2\Big)
\end{equation}
and due to the rank one perturbation of the unit matrix the inverse covariance matrix can be calculated as
\begin{equation}\label{eq:inverse}
  \mathbf{\Sigma}^{-1}_k=\sigma^{-2}I_{P}-\frac{\sigma^{-4}\sigma_{\nabla}^{2}}{1+\sigma^{-2}\sigma_{\nabla}^{2}|\nabla L_n^{(k)}|^2}\nabla L_n^{(k)}\big(\nabla L_n^{(k)}\big)^\top.
\end{equation}
and the probability of the proposal \eqref{eq:proposal2} can thus be evaluated as two vector products.

Note that the directional noise affects the invariant distribution chain. 
Without a M-H step, a correction term to the gradient update, needs to be applied. 
Following the general framework of~\citet{ma2015complete}, this correction would include second derivatives of the loss. 

\subsection{Metropolis-Hastings with \Adam} \label{sec:math}
The change in the proposal distribution is especially effective in combination with momentum variables. 
%
We therefore replace the gradient $\nabla L_n^{(k)}$ in the above construction with an \Adam update step~\citep[Algorithm 1]{Adam2014}. 
The $k$-th step of \Adam updates the momenta $m^{(k+1)}:=\big(m_{1}^{(k+1)},m_{2}^{(k+1)}\big)$ as
\begin{equation}\label{eq:mom_up}
    \begin{aligned}
        &m_{1}^{(k+1)}:=\beta_{1}m_1^{(k)}+(1-\beta_{1})\nabla L_n^{(k)}, \\
        &m_{2}^{(k+1)} := \beta_{2}m_2^{(k)}+(1-\beta_{2})(\nabla L_n^{(k)})^{2},
    \end{aligned}
\end{equation}
where $\beta_{1},\beta_{2}\in[0,1)$ tune the importance of the momenta and the exponent is understood component wise. 
The network parameters are then updated as
\begin{equation*}
    \vartheta^{(k+1)} := \vartheta^{(k)}-u_{k}(m^{(k+1)}),
\end{equation*}
with
\begin{equation*}
    u_{k}(m^{(k+1)}):=\gamma\frac{m_1^{(k+1)}}{1-\beta_{1}^{k+1}} 
    \Big/\Big(\Big(\frac{\big\vert m_2^{(k+1)}\big\vert_{\bullet}}{1-\beta_{2}^{k+1}}\Big)^{1/2}+\delta\Big) 
\end{equation*}
Again, exponents and division are understood component wise, $\vert\cdot\vert_{\bullet}$ denotes the entry-wise absolute value and a small constant $\delta>0$ prevents numerical nuisance.


Due to the dependence on the momenta, 
the augmented chain $(\vartheta^{(k)},m^{(k)})_{k\ge1}$ has to be
considered. 
The augmentation recovers the Markovian nature of the chain and allows us to calculate the acceptance probabilities, as in \eqref{eq:acc_prob}, based on the proposal distributions $q(\vartheta^{(k)}|\tau^{(k)},m^{(k+1)})$ and  $q(\tau^{(k)}|\vartheta^{(k)},m^{(k+1)})$ using the same momenta $m^{(k+1)}$, see \algref{alg:MALAdam}. 
\RestyleAlgo{ruled}
\begin{algorithm}[t]
    \DontPrintSemicolon
\SetKwInOut{Input}{Input}
\Input{empirical~loss~function~\mbox{$L_n(\vartheta)$}, 
 starting~position~$\vartheta^{(0)}$,
 inverse~temperature~\mbox{$\lambda>0$}, learning~rate~\mbox{$\gamma>0$}, 
momenta~parameters~\mbox{$\beta_{1},\beta_{2}\in[0,1)$}, standard~deviations~\mbox{$\sigma, \sigma_\nabla>0$}.
}

 \BlankLine
 
 \For{$k=0,1,2...$}{
    %
    \textit{Sample parameters around} $\Adam(L_n(\vartheta^{(k)}))$\textit{:}\\
    $m^{(k+1)} = (m_{1}^{(k+1)},m_{2}^{(k+1)}) \leftarrow \left(\beta_{1}m_{1}^{(k)}+(1-\beta_{1})\nabla L_n(\vartheta^{(k)}), \beta_{2}m_{2}^{(k)}+(1-\beta_{2})\nabla L_n(\vartheta^{(k)})^2\right)$\;
    $\begin{aligned}
        &\tilde{\vartheta}^{(k+1)}
        \;\leftarrow \vartheta^{(k)}-u_{k}(m^{(k+1)})\\
    \end{aligned}$\;
    $\tau^{(k)} \sim \mathcal{N}\left(\,\tilde{\vartheta}^{(k+1)}, 
    \mathbf{\Sigma}\left(u_{k}(m^{(k+1)})\right)\right)$ \;
    \, \\
    \textit{Metropolis-Hastings correction:}\;
    $\tilde{\tau}^{(k+1)} \leftarrow \tau^{(k)}-u_{k}(m^{(k+1)})$\;
    $\alpha_{k}(\tau^{(k)}|\,\vartheta^{(k)},m^{(k+1)}) =  \mathbf{1}_\Omega(\tau^{(k)})
    \cdot\frac{\exp \left(-\temp L_n(\tau^{(k)})\right)}{\exp \left(-\temp L_n(\vartheta^{(k)})\right)}
    \frac{\varphi_{\tilde{\tau}^{(k+1)},
    \mathbf{\Sigma}\left(u_{k}(m^{(k+1)})\right)}
    \left(\vartheta^{(k)}\right)}
    {\varphi_{\tilde{\vartheta}^{(k+1)}, 
    \mathbf{\Sigma}\left(u_{k}(m^{(k+1)})\right)}
    \left(\tau^{(k)}\right)}$\;
    $a\sim \mathrm{uniform(0,1)}$\;
    \eIf{$a \leq \alpha_{k}(\tau^{(k)}|\,\vartheta^{(k)},m^{(k+1)})$}    {$\vartheta^{(k+1)} \;\leftarrow \tau^{(k)}$\;
    }
    {$\vartheta^{(k+1)} \;\leftarrow \vartheta^{(k)}$\;
    }
 }
\caption{\AdamMCMC\label{alg:MALAdam}}
\end{algorithm}
Therein, we denote the probability density function of the $P$-dimensional normal distribution $\mathcal N(\mu,\mathbf{\Sigma})$ by $\varphi_{\mu,\mathbf{\Sigma}}$. 
If the covariance matrix is diagonal, that is $\mathbf{\Sigma}=\sigma^2 I_P$ for some $\sigma>0$, we abbreviate $\varphi_{\mu,\sigma^2}=\varphi_{\mu,\mathbf{\Sigma}}$. 

The proposal for the network parameters is distributed according to
\begin{equation}
q_{1,k}(\tau^{(k)}|\vartheta^{(k)},m^{(k+1)})=\varphi_{\vartheta^{(k)}-u_{k}(m^{(k+1)}),\mathbf{\Sigma}_{k}}(\tau^{(k)})\label{eq:q1}
\end{equation}
with covariance matrix 
$$\mathbf{\Sigma}_k=\sigma^2 I_{P}+\sigma^2_\nabla u_{k}(m^{(k+1)})u_{k}(m^{(k+1)})^\top.$$
Although we do not randomize the moments in practice, for theoretical considerations we impose a momentum update distribution given by 
\begin{equation}
        q_{2}(m^{(k+1)}|\vartheta^{(k)},m^{(k)})=
        \prod_{l=1}^{2}\varphi_{\beta_{l}m_{l}^{(k)}+(1-\beta_{l})\nabla L_n(\vartheta^{(k)})^{l},\rho_{l}^{2}}(m_{l}^{(k+1)})\label{eq:q2}
\end{equation}
with small noise levels $\rho_{1},\rho_{2}>0$.
This approximation is in line with an additional stochastic error that occurs if the gradient is replaced by a stochastic gradient for Section~\ref{sec:example} and similar approximations are applied for example by \citet{chen2014stochastic}.
The acceptance probabilities are then given by
\begin{equation}
    \begin{aligned}
        \label{eq:acc_adam}
        \alpha_{k}(\tau^{(k)}|\vartheta^{(k)},m^{(k+1)}) = 
        1\wedge\Big(\frac{p_\lambda(\tau^{(k)}|\mathcal D_n)}{p_\lambda(\vartheta^{(k)}|\mathcal D_n)}\frac{q_{1,k}(\vartheta^{(k)}|\tau^{(k)},m^{(k+1)})}{q_{1,k}(\tau^{(k)}|\vartheta^{(k)},m^{(k+1)})}C(\vartheta^{(k)},\tau^{(k)})\Big)
    \end{aligned}
\end{equation}
(setting $0/0=0$ if $p_\lambda(\vartheta^{(k)}|\mathcal D_n)=0$) with a correction term 
\begin{align*}
C(\vartheta^{(k)},\tau^{(k)}) =
\exp{\Big(-\sum_{l=1}^{2}\frac{|m_{l}^{(k+1)}-\nabla L_n(\tau^{(k)})^{l}|^{2}}{2\rho_l^{2}/(1-\beta_l^2)}}
\quad+\sum_{l=1}^{2}\frac{|m_{l}^{(k+1)}-\nabla L_n(\vartheta^{(k)})^{l}|^{2}}{2\rho_l^{2}/(1-\beta_l^2)}\Big).
\end{align*}
Note that $\beta_{1}$ and $\beta_{2}$ are chosen close to $1$ and after a burn-in time the gradients $\nabla L_n(\vartheta^{(k)})^l$ and $\nabla L_n(\tau^{(k)})^l$ will be small and close to their long-term average $m^{(k)}_l$. While our theoretical results rely on the explicit form of $C(\theta^{(k)},\tau^{(k)})$, the correction term can be well approximated by $1$ in the sampling stage and in our algorithm \AdamMCMC we simply set $C(\vartheta^{(k)},\tau^{(k)})=1$.

The following theorem verifies that the \Adam based Metropolis-Hastings algorithm indeed admits the desired invariant distribution.
\begin{thm}\label{thm:invariant_dist}
For arbitrary proposal
distributions $q_{1,k}(\tau^{(k)}|\vartheta^{(k)},m^{(k+1)})$ and $q_2$ from \eqref{eq:q2} the Markov
chain $(\vartheta^{(k)},m^{(k)})_{k\ge1}$ admits the invariant distribution
$$ 
\begin{aligned}
    f(\vartheta,m)=p_{\lambda}(\vartheta|\mathcal{D}_{n}) \mathrm{\varphi}_{\nabla L_n(\vartheta),\rho_1^{2}/(1-\beta_1^2)}(m_{1})
    \cdot\mathrm{\varphi}_{\nabla L_n(\vartheta)^{2},\rho_2^{2}/(1-\beta_2^2)}(m_{2}).
\end{aligned} $$
In particular, the marginal distribution of $f(\vartheta,m)$ in $\vartheta$
is the Gibbs posterior distribution $p_{\lambda}(\cdot|\mathcal{D}_{n})$.
\end{thm}

Moreover, the following result shows a good approximation of the Gibbs distribution in the presence of a sufficiently small momentum. The special case $\beta=0$ corresponds to a Metropolis-Hastings-within-Gibbs algorithm where we obtain convergence to the invariant distribution with the typical geometric rate, cf. \cite{JonesEtAl2014}:

\begin{thm}
\label{thm:convergence-1} Suppose that $L_n(\vartheta)$ and $\nabla L_n(\vartheta)$ are uniformly bounded for $\vartheta\in\Omega$. 
Choose
$q_{1,k}$ and $q_2$ from \eqref{eq:q1} and \eqref{eq:q2}, respectively. Further, let $m^{(0)}\sim \mathcal{N}\big(\nabla L_{n}(\vartheta^{(0)}),\rho_{1}^{2}/(1-\beta_{1}^{2})I_P\big)\otimes\mathcal{N}\big(\nabla L_{n}(\vartheta^{(0)})^{2},\rho_{2}^{2}/(1-\beta_{2}^{2})I_P\big)$, where $\vartheta^{(0)}$ is drawn from an arbitrary distribution with bounded density $f^{(0)}(\vartheta)$ and support $\Omega$.
Then, the total variation distance of the distribution of $\vartheta^{(k)}$ to the Gibbs posterior $p_{\lambda}(\cdot|\mathcal{D}_{n})$ satisfies:
\[
\operatorname{TV}(\mathbb{P}^{\vartheta^{(k)}},\Pi_{\lambda}(\cdot|\mathcal{D}_{n}))\lesssim(1-a)^{k}+ b \beta.
\]
for some $a\in(0,1),b>0$ and $\beta=\beta_1\lor \beta_2\coloneqq \max\{\beta_1,\beta_2\}$.
\end{thm}
Note that the geometric decay suffers from an exponential decrease of $a$ with $P$ as typically observed in the convergence analysis of Metropolis-Hastings algorithms.

\section{Numerical Experiments}\label{sec:example}

We determine the effects of the algorithm parameters $\lambda$, $\sigma$, $\sigma_\Delta$ and ${\beta_1, \beta_2}$ on the generated ensemble of network weights for a high-dimensional classification task from particle physics. 
In a parallel application on neural posterior estimation using continuous normalizing flows \citep{Bieringer2024surrogates}, we have already found great improvements in indicating out-of-distribution input over the commonly used variational inference-based network weight posterior approximation \citep{bbb2015}. 

In neural network training code, the  gradient calculation and optimizer step can be easily exchanged by an \AdamMCMC step.
As the algorithm extends a single \Adam update step, the M-H step constitutes the main computational bottleneck.
Besides the calculation of the loss-values, it only applies highly parallelisable subtractions and vector products.

\subsection{Top-Tagging and ParticleNet}\label{ssec:intro_tt}

\begin{figure*}[t]
    \centering
    \includegraphics[width=\textwidth]{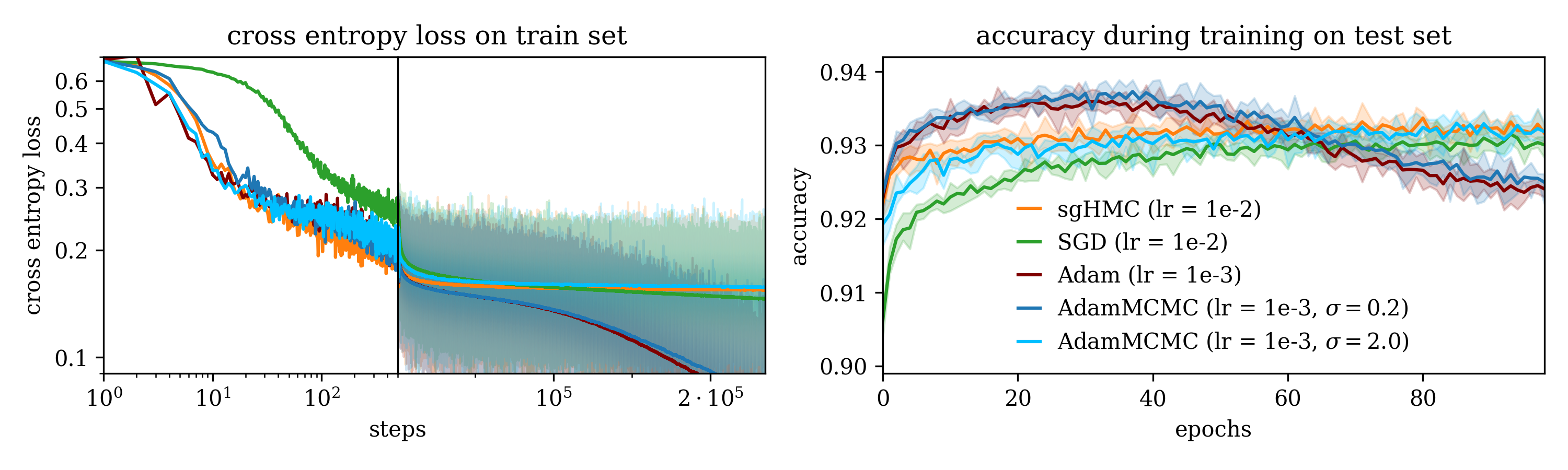}
    \caption{Optimization speed of \AdamMCMC in comparison to \Adam, SGD and sgHMC for the Top-tagging (binary classification) task.
    We report the mean curves over $5$ independent runs for the highest learning rate that allows stable results from a grid search.
    Left: Development of the cross entropy loss on the training set during running of the chain/training. 
    To increase the readability, we show $\log$-scaling for the first $500$ steps and linear scaling as well as the a moving-average over $2400$ steps for the remaining $\approx 230$k steps.
    Right: Accuracy on the test set ($400$k jets) over training epochs ($2400$ steps per epoch).
    While \AdamMCMC ($\sigma = 0.2$) closely resamples the behavior of \Adam including overfitting, the samples generated with \AdamMCMC ($\sigma = 2.0$) show no signs of overfitting and a similar optimization performance as sgHMC.
    The error bands show the $\min$-$\max$ evelope of the $5$ runs.
    }
    \label{fig:loss}
\end{figure*}

At the Large Hadron Collider (LHC), products of particle collisions are measured using calorimeters. 
In these, the generated particles produce a spray of daughter particles each depositing energy in the calorimeter cells.
The cascade of measured energy depositions is usually referred to as a particle shower.
During evaluation, these measured showers are grouped into jets originating in one initial particle.
To reconstruct the correct scattering process in the particle collisions, the reconstructed jets need to be assigned to the correct initial partons.

One particularly useful tool in the investigation of the Higgs particle is a classification of jets originating from Top-quarks from their background originating from lighter quarks (QCD).  
%
To ensure a fair comparison between the multiple efforts within the high-energy physics community, \citet{Kasieczka2019TopLandscape} introduce the TopLandscape dataset.
It contains $600$k top and background jets each for training.

While the Particle Transformer architecture \citep{Qu2022ParT} currently reports the best accuracy and rejection rates, 
we choose the commonly used and more parameter-efficient ParticleNet architecture \citep{Qu2019ParticleNet} for our evaluation of \AdamMCMC. 
ParticleNet constructs a graph from the per-jet point cloud of constituents by connecting each constituent to its $k=16$ nearest neighbours in physical space. 
It does so for the $128$ particles with highest transverse momentum.
We follow the original architecture and apply three layers of edge convolutions \citep{wang2019EdgeConv} to the graph, dynamically recalculating the neighbours at the beginning of every edge convolution block and transforming the features of the graph with a three-layered perceptron based on its neighbours. 
The graph layers are followed by a global average pooling layer on the channels of the edge convolutions. 
After this, two fully connected layers, the first featuring additional dropout of $0.1$ and ReLU activation, and a softmax function are applied.
In total the employed ParticleNet uses $P = 366160$ parameters on an input dimension of $128$ points of $2$ coordinates and $7$ input features.

The classification is trained using a cross entropy loss, that is the sum of the categorical-$\log$-liklihoods per event.
Due to the size of network and dataset, we have to employ stochastic approximations of the loss and its gradient for update and correction steps.
While the stochasticity in the update steps is corrected by the M-H correction, the stochasticity of the M-H step allows a remaining bias to the invariant distribution.
In Section~\ref{ssec:stochMH}, we gauge the effects of this approximation numerically.

Originally, ParticleNet is trained using \texttt{AdamW}\xspace and weight decay. 
To focus on the transition from \Adam to \AdamMCMC, we omit these technicalities and train with a constant learning rate of $1\times10^{-3}$ and $\beta_1 = \beta_2 = 0.99$ for $100$ epochs ($2400$ batches of size $512$ each).
This barely slows down convergence and reports comparable accuracy values to the original training schedule reported by \citet{Qu2019ParticleNet}.

For an initial setup with $\lambda = 1$, $\sigma = 0.2$ and $\sigma_\Delta = P/100 =  3661.6$ shown in Figure~\ref{fig:loss}, we find the sampling algorithm follows its deterministic counterpart closely for the full optimization.
The performance on the $400$k test training set indicates overfitting of the model for both algorithms.
This is a clear indication for running \AdamMCMC with too little noise, thereby prohibiting any parameter space exploration.
We thus increase $\sigma$ to $2.0$, which is the lowest noise setting that allows sufficient parameter space exploration to prevent overfitted samples.

With the higher noise level, \AdamMCMC converges at similar speed as sgHMC.
Both outperform simple stochastic gradient descent (SGD) optimization, which acts as a benchmark for stochastic gradient Langevin Dynamics~\citep{Welling2011bayesianSGLD} based chains, such as MALA.

\subsection{Noise, Directional Noise and Momentum}\label{ssec:dir_noise}

\begin{figure*}
    \centering
    \includegraphics[width=\textwidth]{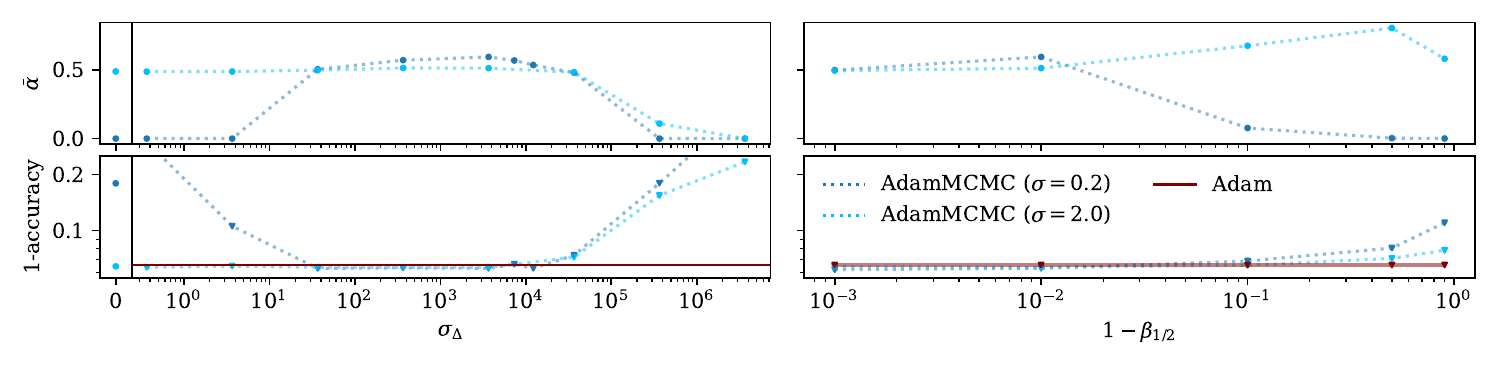}
    \caption{Dependence of the mean acceptance rate (upper) and the accuracy of the posterior mean prediction on test data (lower) on the momentum parameters of the proposal distribution $\sigma_\Delta$ (left) and \Adam optimizer $\beta_{1} = \beta_{2}$ (right).
    For low noise runs, we find a strong dependency of the algorithm efficiency on sufficiently large momentum terms.
    At higher noise, this dependence is reduced due to increased width of the proposal distribution and corresponding higher probability of the backwards direction.}
    \label{fig:02a}
\end{figure*}

\begin{figure}[b]
    \centering
    \includegraphics[width=.5\linewidth]{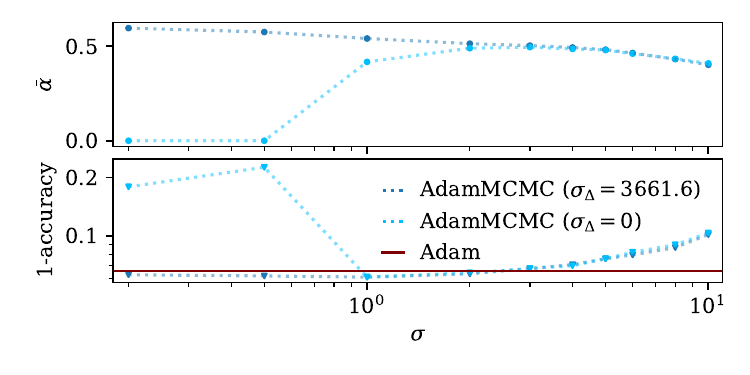}
    \caption{Dependence of the mean acceptance rate (upper) and the accuracy of the posterior mean prediction on test data (lower) on the width of the proposal distribution $\sigma$.
    Without directional noise (light blue), the algorithms efficiency is strongly dependent on a correct choice of $\sigma$.
    Applying an prolate proposal distribution however allows the algorithm to approach the deterministic optimization in the limit of low $\sigma$.
    Noise can then be added to guarantee sufficient parameter space exploration and achieve well-calibrated uncertainties.
    }
    \label{fig:02b}
\end{figure}

The key novelty of the proposed algorithm is the combination of an prolate proposal distribution with momentum-based optimization. 
To gauge the effect of these changes, we observe the dependence of the mean acceptance rate during sampling, as well as the accuracy of the ensemble drawn from the approximate posterior, on both the noise in update direction $\sigma_\Delta$ and the momentum parameters $\beta_1$ and $\beta_2$ in Figure~\ref{fig:02a}.
For easier evaluation, we choose $\beta_1=\beta_2$.
To ensure sampling from a converged chain and approximate independence of the samples, we use a burn-in time of $b=48\cdot2400$ steps and a gap length of $c=5\cdot2400$ steps, that is five epochs, in the following.
From the $100$ epoch runs, we thus generate $N=10$ weight samples.

We find a strong dependence on both, the directional noise and the momentum parameters, when running the algorithm with low noise of $\sigma=0.2$.
As expected for low directional noise ($\sigma_\Delta < 100$) and low momentum ($\beta_{1/2} < 0.99$), the sampling breaks down due to low acceptance rates.
The low acceptance probabilities originate from the low probability of the backwards direction  in~\eqref{eq:acc_adam}.
The accuracy drops accordingly.
Using higher directional noise and momentum increases the acceptance probabilities by aligning the proposal distribution with the optimization step.

For higher noise, $\sigma=2.0$, the same dependence cannot be observed. 
Even without directional noise or momentum, stable sampling at good performance can be achieved due to the sufficient spread of the proposal distribution.

High and low noise runs show a strong decrease in acceptance and performance when going to very high directional noise values ($\sigma_\Delta > 10^5$).
This decrease is caused by the low likelihood values, that is high NLL-loss values, of network parameters sampled with high variance and thus far away from the currently explored loss minimum. 
Similar scaling can be observed at very high overall noise ($\sigma>10$).

Up to this point, the improvements of combining momentum-based optimization with an prolate proposal distribution over sgMALA appear only at low $\sigma$.
In this region, \AdamMCMC is prone to overfitting due to limited parameter space exploration.
Higher values of $\sigma$, in which the effect of the prolate proposal is limited, are to be used regardless.
However, in application the extended algorithm is more well-behaved.
To show this, we run the sampling chain with and without the adapted proposal for an array of noise values.
Figure~\ref{fig:02b} shows the scaling of the mean acceptance and test accuracy for both.

Running a chain at low noise without directional noise leads to diminishing acceptance probabilities, as does running at high noise.
From applying \AdamMCMC to multiple tasks ~\citep{Bieringer2024_bayesiamplify, Bieringer2024surrogates}, we find the range of functional $\sigma$-values can vary strongly between different applications, data set sizes and parameter dimensionalities.
In a hyperparameter search, it is thus unclear in which direction the noise parameter has to be altered to achieve efficient sampling.
When including the directional noise however, low noise parameters result in increasing acceptance and close resemblance between optimizer and MCMC.
Starting from this parameter space interval of semi-deterministic optimization, a practitioner can simply increase the noise value until the mode exploration capabilities are sufficient, overfitting is avoided and the desired uncertainty calibration is achieved.
This renders fine-tuned learning rate schedules as required for sgHMC unnecessary.

\subsection{Error Estimation}

The noise parameter $\sigma$ not only gives a handle on the fitting, but also determines the uncertainty in the predictions.
A higher noise value will lead to larger differences between the different weight samples.
It thus allows us to adjust the uncertainty quantification to calibrate the predictions.

\begin{figure*}[t]
    \centering
    \includegraphics[width=\textwidth]{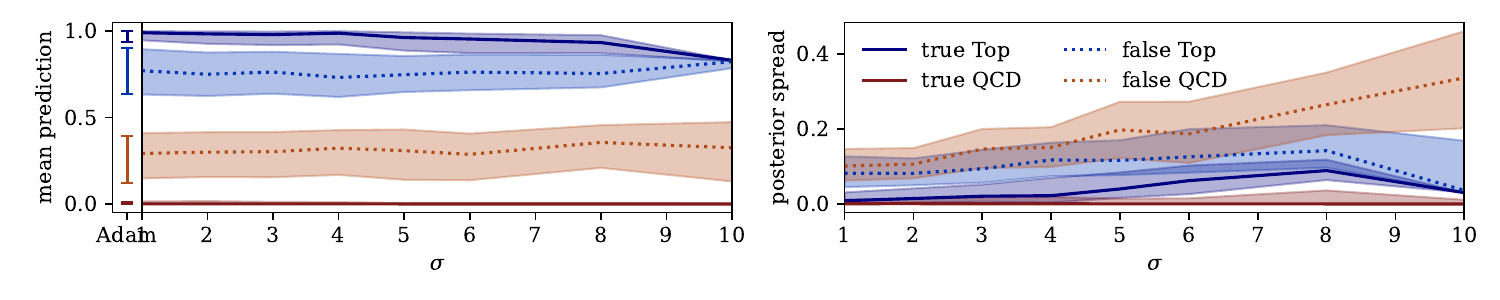}
    \caption{Scaling of the posterior mean prediction (left) and 
    posterior spread (right) with the proposal distribution width $\sigma$.
    The posterior spread is calculated as the difference between the 75\%- and 25\%-quantile of the prediction for $10$ posterior samples.
    We report individual results for both classes and true and false assignments for a test set of $400$k jets.
    The center line shows the median of the jets in the according category and the envelopes depict the 75\%- and 25\%-quantile on the data.
    We find a slight decrease classification performance for rising $\sigma$ for a large section of the scanned space, while for low noise values the performance of an \Adam optimization is closely reproduced.
    The uncertainty prediction increases with increasing $\sigma$ starting out at an already significant level for $\sigma = 1$.
    }
    \label{fig:03}
\end{figure*}

Figure~\ref{fig:03} shows the posterior mean prediction and the posterior spread, as an estimate of the epistemic uncertainty, for different values of $\sigma$.
We refrain from evaluating at $\sigma<1$, due to the previously reported issues with limited mode exploration at low noise.
The class probability predictions are calculated for the $400$k point test set and split into true and false class assignments for both classes, Top- and QCD-jets.

For low noise, we find the posterior mean predictions in the left panel align very well with class predictions from stochastic optimization. 
Increasing the sampling noise to $\sigma > 7$ leads to the decrease in classification power, that has already been observed in Figure~\ref{fig:02b}.
Distinguishing between the two classes allows us attribute this trend to a decrease in predicted probability of Top-jets.

While the performance slightly decreases between $\sigma=1$ and $7$, we find the posterior spread, determined as the distance between the 75\%- and 25\%-quantile of the weight samples, steadily rises.
This is most prominent in the falsely assigned classes.
For these, a significant uncertainty is reported at low noise already.

When running the classification on out-of-distribution data from the more comprehensive JetClass dataset~\citep{qu_2022_6619768_jetclass}, we find a largely increased posterior spread.
The increase of the uncertainty with increasing noise is reproduced analogously to the in-distribution sample.

From the tempered Gibbs posterior~\eqref{eq:gibbs}, we would expect a similar effect from the inverse temperature $\lambda$.
We have varied the inverse temperature within $\lambda \in [10^{-4},10^3]$ and did not find a strong dependence of the uncertainty prediction on this parameter.
Very low values will however lead to strongly suppressed acceptance rates and a corresponding loss of classification performance.

\subsection{Stochastic Metropolis-Hastings}\label{ssec:stochMH}

\begin{figure*}[t]
    \centering
    \includegraphics[width=\textwidth]{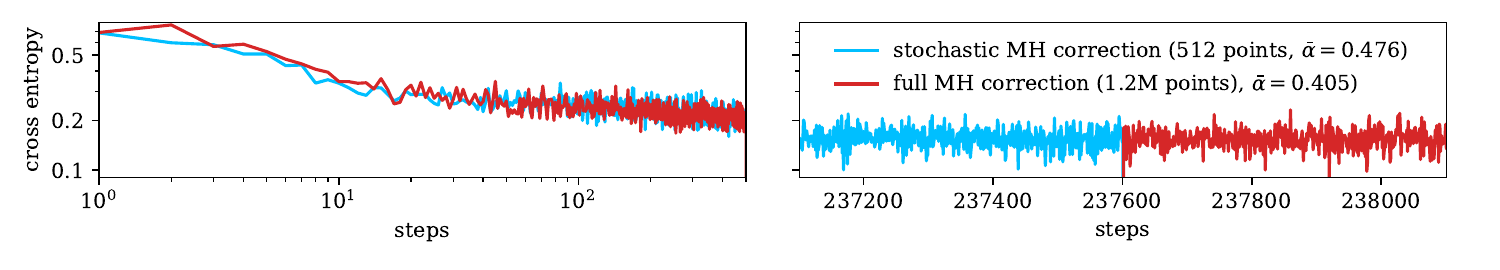}
    \caption{Comparing the loss development for \AdamMCMC algorithms employing a full Metropolis-Hastings correction, as well as a stochastic approximation thereof from random initialization (left) and the end of the stochastically corrected chain (right).
    Both chains are calculated at the same hyperparameters, that is $\sigma=2.0$, $\sigma_\Delta = 3661.6. \lambda = 1, \beta_{1/2}=0.99$ and a learning rate of $10^{-3}$. 
    We find no significant differences in the dynamics of the chain, although the full correction is slightly more selective.
    }
    \label{fig:04}
\end{figure*}

To reduce computational cost for our experiments, we have used a stochastic approximation of the M-H correction.
That is a correction calculated from an unbiased, batch-based estimator of the full loss.
This stochastic correction introduces a bias to the  posterior estimation.

To gauge the difference between an algorithm with a full correction to one using only a batch of data, here $512$ points, we run both algorithms at the same hyperparameter settings.
Due to the immense computational cost of the full correction, we are limited to short chains only.
We thus only evaluate the most interesting regions in Figure~\ref{fig:04}.

Starting from the same random initialization, the differences between both algorithms does not seem to exceed the random fluctuations of the proposal and stochasticity of the batches.
Both algorithms explore the phase space in the same way.

To gauge the sampling after burn-in, we initialize a second chain with full corrections from the end of the chain employing the stochastic approximation.
We find that both, the mean and variance of the loss-values during the chain are virtually the same.
A difference in the mean acceptance probability can however be found.
As expected, the full correction is more restrictive as its stochastic counterpart.

The additional noise introduced by a stochastic correction, can in part be countered by a reduction of  $\sigma$ and $\sigma_\Delta$.
Stochastic M-H corrections that control the introduced bias can be employed for \AdamMCMC, whenever the application requires a strict control of the uncertainties.
While the corrections of \citet{Balan2014stochMH} and \citet{Bardenet014stochMH} rely on subsets of various size to perform sequential hypothesis testing, the minibatch acceptance test of \citet{Seita2018stochMH} ensures detailed balance from fixed size batches with an additive correction variable to a Barker test.
\Citet{Zhang2020_asympt_optim_MH} introduce an exact routine for M-H algorithms on subsamples of data from bounds of the difference in the loss through the update.
Recent proposals \citep{bieringer2023statistical, kawasaki2022data} also propose a correction term to the loss to counteract the batch-wise approximation error of the acceptance probabilities.

\section{Conclusion}\label{sec:conc}


In this report, we proposed a generalization of the Metropolis Adjusted Langevin Algorithm with update steps calculated from \Adam. We suggested a prolate deformation of the proposal distribution to increase the algorithms acceptance rate.
Our construction allows for an efficient calculation of the proposal density which is strictly necessary in order to obtain a computationally feasible algorithm.
We have proven that the resulting algorithm admits the desired Gibbs posterior distribution as invariant distribution. While a general convergence result is left open for further research, we have verified that the Gibbs posterior can be well approximated by the algorithm.

For a classification task for particle physics, we show the algorithm works well with stochastic approximation of loss values and gradients.
\AdamMCMC  recovers the behavior of the underlying stochastic optimization, and thereby improves the robustness of the algorithm, at low injected noise.
Through changing the width of the proposal distribution, it enables the user to adjust the uncertainty quantification starting out from \Adam-like behavior.


\appendix
\section{Proofs}\label{sec:proofs}
In this section, we verify our theoretical contributions. The proof strategies are in line with the literature, see e.g.\,\cite{chauveau2002improving} who have used a step-dependent proposal distribution. However, some technical modifications are necessary to handle noisy momenta.

\subsection{Proof of Theorem~\ref{thm:invariant_dist}}

For brevity, we write $\vartheta=\vartheta^{(k)},\tilde{\vartheta}=\vartheta^{(k+1)}$
and similarly for $m$. The overall transition kernel is
\begin{equation*}
  q_{k}(\tilde \vartheta,\tilde m|\vartheta,m)
  =q_{1,k}(\tilde \vartheta|\vartheta,\tilde m)\alpha(\tilde\vartheta|\vartheta,\tilde m)q_{2}(\tilde m|\vartheta, m)
 +\Big(\int q_{1,k}(\tau|\vartheta,\tilde m)(1-\alpha(\tau|\vartheta,\tilde m))\,\mathrm{d}\tau\Big)q_{2}(\tilde m|\vartheta,m)\,\delta_{\vartheta}(\mathrm{d}\tilde \vartheta),
\end{equation*}
where the first term corresponds to accepting the proposal for $\tilde\vartheta$, the second one rejects the proposal and $\delta_{\vartheta}$ denotes the Dirac measure in $\vartheta$.
By construction we can rewrite
\[
\alpha_{k}(\tilde\vartheta|\vartheta,\tilde m)=1\wedge\frac{f(\tilde\vartheta,\tilde m)q_{1,k}(\vartheta|\tilde\vartheta,\tilde m)}{f(\vartheta,\tilde m)q_{1,k}(\tilde\vartheta|\vartheta,\tilde m)}
\]
and thus the detailed balance equation
\begin{align}
\alpha_{k}(\tilde\vartheta|\vartheta,\tilde m)q_{1,k}(\tilde\vartheta|\vartheta,\tilde m)f(\vartheta,\tilde m) & =\big(f(\tilde\vartheta,\tilde m)q_{1,k}(\vartheta|\tilde\vartheta,\tilde m)\big)\wedge\big(f(\vartheta,\tilde m)q_{1,k}(\tilde\vartheta|\vartheta,\tilde m)\big)\label{eq:detailed_balance}\\
 & =\alpha_{k}(\vartheta|\tilde\vartheta,\tilde m)q_{1,k}(\vartheta|\tilde\vartheta,\tilde m)f(\tilde\vartheta,\tilde m)\nonumber
\end{align}
is satisfied. Setting $s_l^2:= \rho_l^2/(1-\beta_l^2)$ we have $s_l^{2}=\beta_{l}^{2}s_l^{2}+\rho_{l}^{2}$ and thus we deduce for two independent
standard-normal random vectors $Z_{1},Z_{2}\in\R^{P}$ and
any $A=A_{1}\times A_{2}$ with measurable $A_{1},A_{2}\subseteq\R^{P}$ that
\begin{align}
\int_{A}\int q_{2}(\tilde m|\vartheta,m)f(\vartheta,m)\,\mathrm{d}m\,\mathrm{d}\tilde m & =p_{\lambda}(\vartheta|\mathcal{D}_{n})\int_{A}\int q_{2}(\tilde m|\vartheta,m)\prod_{l=1}^{2}\mathrm{\varphi}_{\nabla L_n(\vartheta)^{l},s_l^{2}}(m_{l})\,\mathrm{d}m\,\mathrm{d}\tilde m\nonumber \\
 & =p_{\lambda}(\vartheta|\mathcal{D}_{n})\prod_{l=1}^{2}\int_{A_{l}}\int\varphi_{\beta_{l}m_{l}+(1-\beta_{l})\nabla L_n(\vartheta)^{l},\rho_{l}^{2}}(\tilde m_{l})\varphi_{\nabla L_n(\vartheta)^{l},s_l^{2}}(m_{l})\,\mathrm{d}m_{l}\,\mathrm{d}\tilde m_{l}\nonumber \\
 & =p_{\lambda}(\vartheta|\mathcal{D}_{n})\prod_{l=1}^{2}\mathbb{P}(\beta_{l}(\nabla L_n(\vartheta)^{l}+s_lZ_{1})+(1-\beta_{l})\nabla L_n(\vartheta)^{l}+\rho_{l}Z_{2}\in A_{l})\nonumber \\
 & =p_{\lambda}(\vartheta|\mathcal{D}_{n})\prod_{l=1}^{2}\mathbb{P}(\nabla L_n(\vartheta)^{l}+\sqrt{\beta_{l}^{2}s_l^{2}+\rho_{l}^{2}}Z_{1}\in A_{l})\nonumber \\
 & =\int_{A}f(\vartheta,\tilde m)\,\mathrm{d}\tilde m,\label{eq:q2_invariance}
\end{align}
that is $\int q_{2}(\tilde m|\vartheta,m)f(\vartheta,m)\,\mathrm{d}m=f(\vartheta,\tilde m)$.
Therefore,
\begin{align*}
\int\int_{A}q_{k}(\tilde{\vartheta},\tilde m| & \vartheta,m)f(\vartheta,m)\,\mathrm{d}(\tilde{\vartheta},\tilde m)\,\mathrm{d}(\vartheta,m)\\
= & \int\int\1_{A}(\tilde{\vartheta},\tilde m)q_{1,k}(\tilde{\vartheta}|\vartheta,\tilde m)\alpha_{k}(\tilde{\vartheta}|\vartheta,\tilde m)q_{2}(\tilde m|\vartheta,m)f(\vartheta,m)\,\mathrm{d}(\tilde{\vartheta},\tilde m)\,\mathrm{d}(\vartheta,m)\\
 & \qquad\qquad+\int\int\1_{A}(\vartheta,\tilde m)q_{1,k}(\tilde{\vartheta}|\vartheta,\tilde m)\big(1-\alpha_{k}(\tilde{\vartheta}|\vartheta,\tilde m)\big)q_{2}(\tilde m|\vartheta,m)f(\vartheta,m)\,\mathrm{d}(\tilde{\vartheta},\tilde m)\,\mathrm{d}(\vartheta,m)\\
= & \int\int\1_{A}(\tilde{\vartheta},\tilde m)q_{1,k}(\tilde{\vartheta}|\vartheta,\tilde m)\alpha_{k}(\tilde{\vartheta}|\vartheta,\tilde m)f(\vartheta,\tilde m)\,\mathrm{d}(\tilde{\vartheta},\tilde m)\,\mathrm{d}\vartheta\\
 & \qquad\qquad+\int\int\1_{A}(\vartheta,\tilde m)q_{1,k}(\tilde{\vartheta}|\vartheta,\tilde m)\big(1-\alpha_{k}(\tilde{\vartheta}|\vartheta,\tilde m)\big)f(\vartheta,\tilde m)\,\mathrm{d}(\tilde{\vartheta},\tilde m)\,\mathrm{d}\vartheta\\
 & =\int\int\1_{A}(\tilde{\vartheta},\tilde m)q_{1,k}(\vartheta|\tilde{\vartheta},\tilde m)\alpha_{k}(\vartheta|\tilde{\vartheta},\tilde m)f(\tilde{\vartheta},\tilde m)\,\mathrm{d}(\tilde{\vartheta},\tilde m)\,\mathrm{d}\vartheta\\
 & \qquad\qquad+\int\1_{A}(\vartheta,\tilde m)\int f(\vartheta,\tilde m)\,\mathrm{d}(\vartheta,\tilde m)\\
 & \qquad\qquad-\int\int\1_{A}(\vartheta,\tilde m)q_{1,k}(\tilde{\vartheta}|\vartheta,\tilde m)\alpha_{k}(\tilde{\vartheta}|\vartheta,\tilde m)f(\vartheta,\tilde m)\,\mathrm{d}(\tilde{\vartheta},\tilde m)\,\mathrm{d}\vartheta\\
 & =\int\1_{A}(\vartheta,\tilde m)f(\vartheta,\tilde m)\,\mathrm{d}(\vartheta,\tilde m).
\end{align*}
Hence, if the distribution of $(\vartheta^{(k)},m^{(k)})$ is $f$, then
$(\vartheta^{(k+1)},m^{(k+1)})$ is also distributed according to $f$
which is the claimed stationarity.\qed

\subsection{Proof of Theorem~\ref{thm:convergence-1}}

We denote the joint density of $(\vartheta^{(k)},m^{(k)})$ by $f^{k}(\vartheta,m)$. The density of marginal distribution of $\vartheta^{(k)}$ is denoted by $f^k(\vartheta)$ and, e.g.\ the conditional density of $m^{(k)}$ given $\theta^{(k)}=\theta$ by $f^k(m|\vartheta)$ and similarly for $f$. Throughout let $\tilde{\vartheta},\theta\in\Omega,\tilde m,m=(\tilde m_{1},\tilde m_{2})\in\big(\R^{P}\big)^{2}$. The proof is organized in five steps.

\textit{Step 1:} We show that the relative error $D^k(\tilde\theta,\tilde m):= \frac{f^{k}(\tilde\theta,\tilde m)}{f(\tilde\theta,\tilde m)}-1$ remains bounded. Define 
$$
\tilde f^k(\tilde \theta,\tilde m):= \int f^k(\tilde \theta,m)q_2(\tilde m|\tilde \theta,m)\mathrm dm\qquad\text{and}\qquad \tilde D^{k}(\tilde\theta,\tilde m):= \frac{\tilde f^k(\tilde\theta,\tilde m)}{f(\tilde \theta,\tilde m)}-1.$$
With the convention $0/0=0$ and the momentum adjusted detailed balance condition \eqref{eq:detailed_balance},
we get
\begin{align}
f^{k+1}(\tilde{\vartheta},\tilde m) 
& =\int f^{k}(\vartheta,m)q_{1,k}(\tilde{\vartheta}|\vartheta,\tilde m)\alpha_k(\tilde{\vartheta}|\vartheta,\tilde m)q_{2}(\tilde m|\vartheta,m)\,\mathrm{d}(\vartheta,m)\notag\\
 & \qquad\qquad+\int f^{k}(\tilde{\vartheta},m)q_{1,k}(\vartheta|\tilde{\vartheta},\tilde m)\big(1-\alpha_{k}(\vartheta|\tilde{\vartheta},\tilde m)\big)q_{2}(\tilde m|\tilde \vartheta,m)\,\mathrm{d}(\vartheta,m)\notag\\
& =\int \tilde f^{k}(\vartheta,\tilde m)q_{1,k}(\tilde{\vartheta}|\vartheta,\tilde m)\alpha_k(\tilde{\vartheta}|\vartheta,\tilde m)\,\mathrm{d}\vartheta
+\int \tilde f^{k}(\tilde{\vartheta},\tilde m)q_{1,k}(\vartheta|\tilde{\vartheta},\tilde m)\big(1-\alpha_{k}(\vartheta|\tilde{\vartheta},\tilde m)\big)\,\mathrm{d}\vartheta\notag\\
 & =\tilde f^{k}(\tilde{\vartheta},\tilde m)+\int f(\vartheta,\tilde m)q_{1,k}(\tilde{\vartheta}|\vartheta,\tilde m)\alpha_{k}(\tilde{\vartheta}|\vartheta,\tilde m)\frac{\tilde f^{k}(\vartheta,\tilde m)}{f(\vartheta,\tilde m)}\,\mathrm{d}\vartheta\notag\\
 & \qquad\qquad-\int f(\tilde{\vartheta},\tilde m)q_{1,k}(\vartheta|\tilde{\vartheta},\tilde m)\alpha_{k}(\vartheta|\tilde{\vartheta},\tilde m)\frac{\tilde f^{k}(\tilde{\vartheta},\tilde m)}{f(\tilde{\vartheta},\tilde m)}\,\mathrm{d}\vartheta\notag\\
 & =\tilde f^{k}(\tilde{\vartheta},\tilde m)-\int\Big(\frac{\tilde f^{k}(\tilde{\vartheta},\tilde m)}{f(\tilde{\vartheta},\tilde m)}-\frac{\tilde f^{k}(\vartheta,\tilde m)}{f(\vartheta,\tilde m)}\Big)h_{k}(\vartheta,\tilde{\vartheta},\tilde m)\,\mathrm{d}\vartheta, \label{eq:Step1}
\end{align}
where 
\[
h_k(\vartheta,\tilde \vartheta,\tilde m):= \big(f(\tilde\vartheta,\tilde m)q_{1,k}(\vartheta|\tilde\vartheta,\tilde m)\big)\wedge\big(f(\vartheta,\tilde m)q_{1,k}(\tilde\vartheta|\vartheta,\tilde m)\big).
\]
Setting $Q_k(\tilde \vartheta,\vartheta,\tilde m):= h_k(\vartheta,\tilde\vartheta,\tilde m)/f(\tilde \vartheta,\tilde m)$ this leads to
\begin{align*}
D^{k+1}(\tilde{\vartheta},\tilde m) 
 & =\tilde D^{k}(\tilde{\vartheta},\tilde m)-\int\Big(\frac{\tilde f^{k}(\tilde{\vartheta},\tilde m)}{f(\tilde{\vartheta},\tilde m)}-1\Big)\frac{h_{k}(\vartheta,\tilde{\vartheta},\tilde m)}{f(\tilde{\vartheta},\tilde m)}\,\mathrm{d}\vartheta
 +\int\Big(\frac{\tilde f^{k}(\vartheta,\tilde m)}{f(\vartheta,\tilde m)}-1\Big)\frac{h_{k}(\vartheta,\tilde{\vartheta},\tilde m)}{f(\tilde{\vartheta},\tilde m)}\,\mathrm{d}\vartheta\\
 & =\tilde D^{k}(\tilde{\vartheta},\tilde m)-\int \tilde D^{k}(\tilde{\vartheta},\tilde m)Q_{k}(\tilde{\vartheta},\vartheta,\tilde m)\,\mathrm{d}\vartheta+\int \tilde D^{k}(\vartheta,\tilde m)Q_{k}(\tilde{\vartheta},\vartheta,\tilde m)\,\mathrm{d}\vartheta\\
 & =\tilde D^{k}(\tilde{\vartheta},\tilde m)\Big(1-\int Q_{k}(\tilde{\vartheta},\vartheta,\tilde m)\,\mathrm{d}\vartheta\Big)+\int \tilde D^{k}(\vartheta,\tilde m)Q_{k}(\tilde{\vartheta},\vartheta,\tilde m)\,\mathrm{d}\vartheta.
\end{align*}
Since $\int Q_{k}(\tilde{\vartheta},\vartheta,\tilde m)\,\mathrm{d}\vartheta\le 1$, we conclude
\begin{align}
D^{k+1}(\tilde{\vartheta},\tilde m) & \le \|\tilde D^{k}\|_\infty-\int\big(\|\tilde D^k\|_{\infty}-\tilde D^{k}(\vartheta,\tilde m)\big)Q_{k}(\tilde{\vartheta},\vartheta,\tilde m)\,\mathrm{d}\vartheta\label{eq:firstEstimate}
\le \|\tilde D^{k}\|_\infty.
\end{align}
Since \eqref{eq:q2_invariance} yields
\begin{equation*}
  \|\tilde D^k\|_{\infty}=\sup_{\tilde \theta,\tilde m}\Big\vert\frac{\int (f^k(\tilde \theta,m)-f(\tilde \theta,m))q_2(\tilde m|\tilde \theta,m)\,\mathrm d m}{f(\tilde\theta,\tilde m)}\Big\vert
  =\sup_{\tilde \theta,\tilde m}\Big\vert\frac{\int D^k(\tilde \theta,m)f(\tilde \theta,m)q_2(\tilde m|\tilde \theta,m)\,\mathrm d m}{f(\tilde\theta,\tilde m)}\Big\vert
  \le \|D^k\|_\infty,
\end{equation*}
we obtain for all $k\in\mathbb N$
\[
  \|D^{k+1}\|_\infty\le \|D^k\|_\infty\le \|D^0\|_\infty.
\]

\emph{Step 2:} We now study the relative error of the marginal distribution of $\theta$. To this end we write
\begin{gather*}
  F^{k+1}(\theta):= \frac{\Delta f^{k+1}(\theta)}{f(\theta)},\quad \Delta f^{k+1}(\theta):= f^{k+1}(\theta)-f(\theta),\quad\Delta f^{k+1}(\theta,m):= f^{k+1}(\theta,m)-f(\theta,m),\\
  \quad \Delta \tilde{f}^{k+1}(\theta,m):=\tilde{f}^{k+1}(\theta,m)-f(\theta,m).
\end{gather*} 

To bound $\|F^{k+1}\|_\infty$, we start similarly to Step 1. Using \eqref{eq:Step1}, we have
\begin{align*}
  \Delta f^{k+1}(\tilde \theta,\tilde m)&= \Delta f^k(\tilde \theta)f(\tilde m|\tilde \theta)-\int\Big(\frac{\Delta f^k(\tilde \theta)f(\tilde m|\tilde \theta)}{f(\tilde \theta,\tilde m)}-\frac{\Delta f^k(\theta)f(\tilde m|\theta)}{f(\theta,\tilde m)}\Big)h_k(\theta,\tilde\theta,\tilde m)\,\mathrm d \theta\\
  &\qquad +\big(\Delta\tilde f^k(\tilde \theta,\tilde m)-\Delta f^k(\tilde \theta)f(\tilde m|\tilde \theta)\big)\\
  &\qquad -\int\Big(\frac{\Delta\tilde f^k(\tilde \theta,\tilde m)-\Delta f^k(\tilde \theta)f(\tilde m|\tilde \theta)}{f(\tilde \theta,\tilde m)}-\frac{\Delta\tilde f^k(\theta,\tilde m)-\Delta f^k(\theta)f(\tilde m|\theta)}{f(\theta,\tilde m)}\Big)h_k(\theta,\tilde\theta,\tilde m)\,\mathrm d \theta\\
  &=: T_1(\tilde\theta,\tilde m)+T^{k}_2(\tilde\theta,\tilde m)
\end{align*}
with
\begin{align*}
  T_1(\tilde\theta,\tilde m)&=\Delta f^k(\tilde \theta)f(\tilde m|\tilde \theta)-\int\Big(\frac{\Delta f^k(\tilde \theta)f(\tilde m|\tilde \theta)}{f(\tilde \theta,\tilde m)}-\frac{\Delta f^k(\theta)f(\tilde m|\theta)}{f(\theta,\tilde m)}\Big)h_k(\theta,\tilde\theta,\tilde m)\,\mathrm d\theta\\
  &=\Delta f^k(\tilde \theta)f(\tilde m|\tilde \theta)-\int\Big(\frac{\Delta f^k(\tilde \theta)}{f(\tilde \theta)}-\frac{\Delta f^k(\theta)}{f(\theta)}\Big)h_k(\theta,\tilde\theta,\tilde m)\,\mathrm d \theta.
\end{align*}
Hence,
\begin{equation*}
  F^{k+1}(\tilde\theta)=\frac{1}{f(\tilde\theta)}\int\Delta f^{k+1}(\tilde \theta,\tilde m)\,\mathrm d\tilde m
  =F^{k}(\tilde\theta)-\int\big(F^{k}(\tilde\theta)-F^{k}(\theta)\big) \tilde Q_k(\tilde \theta,\theta)\,\mathrm d\theta+R_k(\tilde\theta)
\end{equation*}
with $\tilde Q_k(\tilde\theta,\theta):=\int \frac{h_k(\theta,\tilde\theta,\tilde m)}{f(\tilde\theta)}\,\mathrm d\tilde m$ and $R_k(\tilde\theta):=\frac{1}{f(\tilde \theta)}\int T^{k}_2(\tilde \theta,\tilde m)\,\mathrm d\tilde m$.

\textit{Step 3:} We first bound the main term in the above display. To this end we verify the momentum adjusted Doeblin-type condition \begin{equation}
\tilde Q_k(\tilde\theta,\theta)\ge af(\theta),\qquad \forall \tilde\vartheta,\vartheta\in\Omega\label{eq:Doeblin_type}
\end{equation}
for some $a>0$. 

It holds that 
\begin{align*}
\tilde{Q}_{k}(\tilde{\vartheta},\vartheta)  =\int\frac{h_{k}(\vartheta,\tilde{\vartheta},\tilde{m})}{f(\tilde{\vartheta})}\,\mathrm{d}\tilde{m}
 & =\int\frac{1}{f(\tilde{\vartheta})}\Big(\big(q_{1,k}(\vartheta|\tilde{\vartheta},\tilde{m}\big)f(\tilde{\vartheta},\tilde{m})\big)\land\big(f(\vartheta,\tilde{m})q_{1,k}(\tilde{\vartheta}|\vartheta,\tilde{m})\big)\Big)\,\mathrm{d}\tilde{m}\\
 & =\int\big(q_{1,k}(\vartheta |\tilde{\vartheta},\tilde{m}\big)f(\tilde{m}|\tilde{\vartheta})\big)\land\big(\tfrac{f(\vartheta)}{f(\tilde{\vartheta})}f(\tilde m|\theta)q_{1,k}(\tilde{\vartheta}|\vartheta,\tilde{m})\big)\,\mathrm{d}\tilde{m}.
\end{align*}
Since $\Omega$ is bounded, $\det(\Sigma_{k})$ as calculated in
(\ref{eq:det}) is uniformly bounded from above and away from $0$
and since $\Sigma_{k}^{-1}\le C_{1}$ in the ordering of positive definite
matrices for some $C_{1}>0$, we have 
\begin{align*}
q_{1,k}(\vartheta |\tilde{\vartheta},\tilde{m}\big)f(\tilde{m}|\tilde{\vartheta}) & =\varphi_{\tilde{\vartheta}-u_{k}(\tilde{m}),\Sigma_{k}}(\vartheta)\varphi_{\nabla L_{n}(\tilde{\vartheta}),\rho_{1}^{2}/(1-\beta_{1}^{2})}(\tilde{m}_{1})\varphi_{\nabla L_{n}(\tilde{\vartheta})^{2},\rho_{2}^{2}/(1-\beta_{2}^{2})}(\tilde{m}_{2})\\
 & \gtrsim\exp(-C_{1}\vert u_{k}(\tilde{m})\vert^{2})\varphi_{\nabla L_{n}(\tilde{\vartheta}),\rho_{1}^{2}/(1-\beta_{1}^{2})}(\tilde{m}_{1})\varphi_{\nabla L_{n}(\tilde{\vartheta})^{2},\rho_{2}^{2}/(1-\beta_{2}^{2})}(\tilde{m}_{2})\\
 & \ge\exp(-\tilde{C}_{1}\vert\tilde{m}_{1}\vert^{2})\varphi_{\nabla L_{n}(\tilde{\vartheta}),\rho_{1}^{2}/(1-\beta_{1}^{2})}(\tilde{m}_{1})\varphi_{\nabla L_{n}(\tilde{\vartheta})^{2},\rho_{2}^{2}/(1-\beta_{2}^{2})}(\tilde{m}_{2}).
\end{align*}
With an analogous bound for $\tfrac{f(\vartheta,\tilde{m})}{f(\tilde{\vartheta})}q_{1,k}(\tilde{\vartheta}|\vartheta,\tilde{m})$,
we obtain for some constant $\tilde{C}>0$
\begin{align}
\tilde{Q}_{k}(\tilde{\vartheta},\vartheta) & \gtrsim\int\exp(-\tilde{C}\vert\tilde{m}_{1}\vert^{2})\varphi_{\nabla L_{n}(\tilde{\vartheta}),\rho_{1}^{2}/(1-\beta_{1}^{2})}(\tilde{m}_{1})\varphi_{\nabla L_{n}(\tilde{\vartheta})^{2},\rho_{2}^{2}/(1-\beta_{2}^{2})}(\tilde{m}_{2})\,\mathrm{d}\tilde{m}\nonumber \\
 & =\int\exp(-\tilde{C}\vert\tilde{m}_{1}\vert^{2})\varphi_{\nabla L_{n}(\tilde{\vartheta}),\rho_{1}^{2}/(1-\beta_{1}^{2})}(\tilde{m}_{1})\,\mathrm{d}\tilde{m}_{1}.\label{eq:tilde_Q}
\end{align}
This integral is equal to $\E[\exp(-\vert Z\vert^{2})]$
for $Z=(Z_{1},\dots,Z_{P})\sim\mathcal{N}\big(\mu,\tilde{\sigma}^{2}I_P\big)$
with $\mu=(\mu_{1},\dots,\mu_{P})^{\top}=\sqrt{\tilde{C}}\nabla L_{n}(\tilde{\vartheta})$
and $\tilde{\sigma}^{2}=\tilde{C}\rho_{1}^{2}/(1-\beta_{1}^{2})$. It holds
that
\begin{align}
\E[\exp(-\vert Z\vert^{2})]  =\prod_{i=1}^{P}\E[\exp(-\vert Z_{i}\vert^{2})] & =\frac{1}{(2\pi\tilde{\sigma}^{2})^{P/2}}\prod_{i=1}^{P}\int\exp\big(-\vert z_{i}\vert^{2}\big)\exp\Big(-\frac{1}{2\tilde{\sigma}^{2}}(z_{i}-\mu_{i})^{2}\Big)\,\mathrm{d}z_{i}\nonumber \\
 & =\frac{1}{(2\tilde{\sigma}^{2}+1)^{P/2}}\exp\Big(-\frac{\vert\mu\vert^{2}}{2 \tilde{\sigma}^{2}+1}\Big).\label{eq:exponential_moments}
\end{align}
Noting that $\vert\mu\vert$ is bounded, $\tilde{\sigma}^{2}$ is
bounded and bounded away from $0$ and $f$ is bounded and bounded
away from $0$ on its support, (\ref{eq:tilde_Q}) and (\ref{eq:exponential_moments})
yield (\ref{eq:Doeblin_type}).

Note that $\int \tilde Q_k(\tilde\theta,\theta)\,\mathrm d\theta\le 1$. Therefore, we can conclude from (\ref{eq:Doeblin_type}) that
\begin{align*}
 F^{k+1}(\tilde \theta)
 &=F^{k}(\tilde\theta)\Big(1-\int\tilde Q_k(\tilde \theta,\theta)\,\mathrm d\theta\Big)+\int F^{k}(\theta)\tilde Q_k(\tilde \theta,\theta)\,\mathrm d\theta+R_k(\tilde\theta)\\
 &\le\|F^{k}\|_\infty -\int \big(\|F^k\|_\infty -F^{k}(\theta)\big)\tilde Q_k(\tilde \theta,\theta)\,\mathrm d\theta+R_k(\tilde\theta)\\
 &\le\|F^{k}\|_\infty - a\int \big(\|F^k\|_\infty -F^{k}(\theta)\big) f(\theta)\,\mathrm d\theta+R_k(\tilde\theta)\\
 &=(1-a)\|F^{k}\|_\infty+a\int(f^{k}(\theta)-f(\theta))\,\mathrm{d}\theta+R_k(\tilde\theta)\\
 &=(1-a)\|F^{k-1}\|_\infty+R_k(\tilde\theta).
\end{align*}
With an analogous bound for $-F^{k+1}$ we obtain
\[
  |F^{k+1}(\tilde \theta)|\le (1-a)\|F^{k-1}\|_\infty+|R_k(\tilde\theta)|.
\]

\emph{Step 4:} Finally we bound the remainder $R_k$. Recall that
$\Delta f^k(\theta,m):= f^k(\theta,m)-f(\theta,m)$ and,  similarly, write  $\Delta f^k(m):= f^k(m)-f(m)$. Using \eqref{eq:q2_invariance}, we have
\begin{align}
  \big|\Delta \tilde{f}^k(\theta,\tilde m)-\Delta f^k(\theta)f(\tilde m|\theta)\big|
  &=\Big|\int \Delta f^k(\theta,m)\big(q_2(\tilde m|\theta,m)-f(\tilde m|\theta)\big)\,\mathrm dm\Big|\notag\\
  &=\Big|\int D^k(\theta,m)f(\theta,m)\Phi(\theta,\tilde m,m)\,\mathrm dm\Big|\notag\\
  &\le \|D^k\|_\infty\int f(\theta,m)|\Phi(\theta,\tilde m,m)|\,\mathrm dm\label{eq:Phi}
\end{align}
with
\begin{align*}
  \Phi(\theta,\tilde m,m)&:= q_2(\tilde m|\theta,m)-f(\tilde m|\theta)=\prod_{l=1}^2\phi_{\beta_l,m_l+(1-\beta_l)\nabla L(\theta)^l,\rho_l^2}(\tilde m_l)-\prod_{l=1}^2\phi_{\nabla L(\theta)^l,\rho_l^2/(1-\beta_l^2)}(\tilde m_l).
\end{align*}
Therefore, we can use $h_k(\vartheta,\tilde\vartheta,\tilde m)/f(\vartheta,\tilde m)\le q_{1,k}(\tilde \vartheta|\vartheta,\tilde m)$ to obtain
\begin{align*}
  |R_k(\tilde\theta)|&\le \frac{1}{f(\tilde\theta)}\int\big|\Delta\tilde f^k(\tilde \theta,\tilde m)-\Delta f^{k}(\tilde \theta)f(\tilde m|\tilde \theta)\big|\Big(1+\int\frac{h_k(\theta,\tilde\theta,\tilde m)}{f(\tilde \theta,\tilde m)}\,\mathrm d\theta \Big)\,\mathrm d\tilde m\\
  &\qquad \qquad\qquad+\frac{1}{f(\tilde\theta)}\iint\big|\Delta\tilde f^{k}(\theta,\tilde m)-\Delta f^{k}(\theta)f(\tilde m|\theta)\big| \frac{h_k(\theta,\tilde\theta,\tilde m)}{f(\theta,\tilde m)}\,\mathrm d \theta\,\mathrm d\tilde m\\
  &\le\|D^{k}\|_\infty\Big(\frac{2}{f(\tilde\theta)} \iint f(\tilde \theta,m)|\Phi(\tilde \theta,\tilde m,m)|\,\mathrm d m \,\mathrm d\tilde m\\
  &\qquad\qquad\qquad + \frac{1}{f(\tilde\theta)}\iiint f(\theta,m)|\Phi(\theta,\tilde m,m)|q_{1,k}(\tilde \theta|\theta,\tilde m)\,\mathrm d m\,\mathrm d\tilde m\,\mathrm d\theta\Big)\\
  &=:C(\tilde\theta, \beta)\|D^{k}\|_\infty.
\end{align*}
From the explicit form of $\Phi$ and the boundedness of $f(\tilde\theta)$ away from $0$, we can now derive a constant $b>0$ such that
\begin{equation}
  C(\tilde\theta, \beta)\le b\beta\label{eq:C_beta}
\end{equation}
for all $\tilde \vartheta\in\Omega$.

To this end, we employ Proposition 2.1 in \cite{Devroye2018} to obtain
\begin{align*}
\int\vert\Phi(\tilde{\vartheta},\tilde{m},m)\vert\,\mathrm{d}\tilde{m} & =2\mathrm{TV}\Big(\bigotimes_{l=1}^{2}\mathcal{N}\big(\beta_{l}m_{l}+(1-\beta_{l})\nabla L(\tilde{\vartheta})^{l},\rho_{l}^{2}I_P\big),\bigotimes_{l=1}^{2}\mathcal{N}\big(\nabla L(\tilde{\vartheta})^{l}\big),\rho_{l}^{2}/(1-\beta_{l}^{2})I_P)\big)\Big)\\
 & \le\Big(\sum_{l=1}^{2}\beta_{l}\rho_{l}^{-1}\vert m_{l}-\nabla L_{n}(\tilde{\vartheta})^{l}\vert^2+\sqrt{P}\big(\beta_{l}/(1-\beta_{l}^{2})+\big(\log(1-\beta_{l}^2)\big)\big)\Big)^{1/2}.
\end{align*}
Therefore,
\begin{align*}
\int&\int f(\tilde{\vartheta},m)\vert\Phi(\tilde{\vartheta},\tilde{m},m)\vert\,\mathrm{d}m\,\mathrm{d}\tilde{m}  \\
&=\int f(\tilde{\vartheta},m)\Big(\int\vert\Phi(\tilde{\vartheta},\tilde{m},m)\vert\,\mathrm{d}\tilde{m}\Big)\mathrm{d}m\\
 & \le\sum_{l=1}^{2}\Big(\beta_{l}\rho_{l}^{-1}\int\vert m_{l}-\nabla L_{n}(\tilde{\vartheta})^{l}\vert f(\tilde{\vartheta},m)\,\mathrm{d}m+\sqrt{P}\big(\beta_{l}/(1-\beta_{l}^{2})^{1/2}+\big\vert\log(1-\beta_{l})\big\vert^{1/2}\big)\int f(\tilde{\vartheta},m)\,\mathrm{d}m\Big)\\
 & =f(\tilde{\vartheta})\sum_{l=1}^{2}\Big(\beta_{l}\rho_{l}^{-1}\int\vert m_{l}-\nabla L_{n}(\tilde{\vartheta})^{l}\vert\varphi_{\nabla L_{n}(\tilde{\vartheta}),\rho_{1}^{2}/(1-\beta_{1}^{2})}(m_{1})\varphi_{\nabla L_{n}(\tilde{\vartheta})^{2},\rho_{2}^{2}/(1-\beta_{2}^{2})}(m_{2})\\
 &\qquad\qquad\qquad\qquad\qquad\qquad+\sqrt{P}\big(\beta_{l}/(1-\beta_{l}^{2})^{1/2}+\big\vert\log(1-\beta_{l})\big\vert^{1/2}\big)\Big)\\
 & \le f(\tilde{\vartheta})\sum_{l=1}^{2}\Big(\beta_{l}\rho_{l}^{-1}\int\vert m_{l}-\nabla L_{n}(\tilde{\vartheta})^{l}\vert\varphi_{\nabla L_{n}(\tilde{\vartheta})^{l},\rho_{l}^{2}/(1-\beta_{l}^{2})}(m_{l})\,\mathrm{d}m_{l}\\
 &\qquad\qquad\qquad\qquad\qquad\qquad+\sqrt{P}\big(\beta_{l}/(1-\beta_{l}^{2})^{1/2}+\big\vert\log(1-\beta_{l})\big\vert^{1/2}\big)\Big).
\end{align*}
Note that for fixed $l=1,2$, the integral term in the last display is
equal to $\E[\vert Z\vert]$ with $Z\sim\mathcal{N}\big(0,(\rho_{l}^{2}/(1-\beta_{l}^{2})I_{P}\big)$.
It holds that
\[
\E[\vert Z\vert]\le\E[\vert Z\vert_{1}]\le\frac{P\rho_{l}}{(1-\beta_{l}^{2})^{1/2}}.
\]
Hence,
\begin{align}
\int\int f(\tilde{\vartheta},m)\vert\Phi(\tilde{\vartheta},\tilde{m},m)\vert\,\mathrm{d}m\,\mathrm{d}\tilde{m} & \le f(\tilde{\vartheta})\sum_{l=1}^{2}\Big(\frac{P\beta_{l}}{(1-\beta_{l}^{2})^{1/2}}+\sqrt{P}\big(\beta_{l}/(1-\beta_{l}^{2})^{1/2}+\big\vert\log(1-\beta_{l}^{2})\big\vert^{1/2}\big)\Big)\nonumber \\
 & =:f(\tilde{\vartheta})C(\beta).\label{eq:Phi}
\end{align}
Further, note that $q_{1,k}(\tilde{\vartheta}|\vartheta,\tilde{m})\le(2\pi\sigma)^{-P/2}.$
Similarly to (\ref{eq:Phi}), we have 
\begin{align*}
\int\int\int f(\vartheta,m)\vert\Phi(\vartheta,\tilde{m},m)\vert q_{1,k}(\tilde{\vartheta}|\vartheta,\tilde{m})\,\mathrm{d}m\,\mathrm{d}\tilde{m}\,\mathrm{d}\vartheta & \le(2\pi\sigma)^{-P/2}\int\Big(\int f(\vartheta,m)\int\vert\Phi(\vartheta,\tilde{m},m)\vert\,\mathrm{d}\tilde{m}\,\mathrm{d}m\Big)\,\mathrm{d}\vartheta\\
 & =(2\pi\sigma)^{-P/2}C(\beta)\int f(\vartheta)\,\mathrm{d}\vartheta\\
 & =(2\pi\sigma)^{-P/2}C(\beta).
\end{align*}
Overall, (\ref{eq:C_beta}) is verified, since $f$ is bounded away from $0$ on its support and $C(\beta)\le b'\beta $ for some $b'>0$.

\emph{Step 5:} Putting everything together, we obtain
\begin{align*}
  \|F^{k+1}\|_\infty \le (1- a)\|F^{k}\|_\infty+b\|D^{0}\|_\infty \beta
  &\le (1- a)^{k+1}\|F^{0}\|_\infty+b\|D^{0}\|_\infty\beta \sum_{l=0}^{k}(1-a)^l\\
  &\le (1- a)^{k+1}\|F^{0}\|_\infty+\frac{b}{a}\beta \|D^{0}\|_\infty.
\end{align*}
Since our choice of the distribution for initializing the chain implies that $F^0,D^0$ are bounded, the proof is complete.
\qed

\section*{Acknowledgments}
SB is supported by the Helmholtz Information and Data Science Schools via DASHH (Data Science in Hamburg - HELMHOLTZ Graduate School for the Structure of Matter) with the grant HIDSS-0002.
SB and GK acknowledge support by the Deutsche Forschungsgemeinschaft (DFG) under Germany’s Excellence Strategy – EXC 2121  Quantum Universe – 390833306.
This research was supported in part through the Maxwell computational resources operated at Deutsches Elektronen-Synchrotron DESY, Hamburg, Germany. 
MS and MT acknowledge support by the DFG through project TR 1349/3-1.

\section*{Code}
The code for running \AdamMCMC in general, as well es the presented experiments is provided at \url{https://github.com/sbieringer/AdamMCMC}

\bibliography{arxiv}
\bibliographystyle{apalike2}

\end{document}